\documentclass[letterpaper]{article} 
\usepackage{aaai24}  
\usepackage{times}  
\usepackage{helvet}  
\usepackage{courier}  
\usepackage[hyphens]{url}  
\usepackage{graphicx} 
\usepackage{natbib}  
\usepackage{caption} 
\usepackage{algorithm}
\usepackage{algorithmic}

\usepackage{amsmath,amssymb}
\usepackage{multirow}
\usepackage{color}
\usepackage{xcolor}
\usepackage{colortbl}
\definecolor{mygray}{gray}{.9}

\newcommand{\ourmodel}{\textit{DI-V2X}}
\newcommand{\eg}{\textit{e.g.}}
\newcommand{\ie}{\textit{i.e.}}

\usepackage{newfloat}
\usepackage{listings}
\DeclareCaptionStyle{ruled}{labelfont=normalfont,labelsep=colon,strut=off} 
\floatstyle{ruled}
\newfloat{listing}{tb}{lst}{}
\floatname{listing}{Listing}
\title{\textit{DI-V2X}: Learning Domain-Invariant Representation for Vehicle-Infrastructure Collaborative 3D Object Detection}
\author{
    Xiang Li\textsuperscript{\rm 1}\equalcontrib,
    Junbo Yin\textsuperscript{\rm 1}\equalcontrib, 
    Wei Li\textsuperscript{\rm 2}, 
    Cheng-Zhong Xu\textsuperscript{\rm 3}, 
    Ruigang Yang\textsuperscript{\rm 2}, 
    Jianbing Shen\textsuperscript{\rm 3}\thanks{Corresponding author (jianbingshen@um.edu.mo). Work supported by the FDCT grants 0154/2022/A3, 0102/2023/RIA2, and SKLIOTSC(UM)-2021-2023, the MYRG-CRG2022-00013-IOTSC-ICI grant and the SRG2022-00023-IOTSC grant.}
}
\affiliations{
    \textsuperscript{\rm 1}School of Computer Science, Beijing Institute of Technology\\
    \textsuperscript{\rm 2}Inceptio \\
    \textsuperscript{\rm 3}SKL-IOTSC, CIS, University of Macau

    \{lixianggoing, yinjunbocn\}@gmail.com
}

\usepackage{bibentry}

\begin{document}

\maketitle

\begin{abstract}

Vehicle-to-Everything (V2X) collaborative perception has recently gained significant attention due to its capability to enhance scene understanding by integrating information from various agents, \textit{e.g.}, vehicles, and infrastructure. However, current works often treat the information from each agent equally, ignoring the inherent domain gap caused by the utilization of different LiDAR sensors of each agent, thus leading to suboptimal performance. In this paper, we propose \textbf{\textit{DI-V2X}}, that aims to learn \textbf{\textit{D}}omain-\textbf{\textit{I}}nvariant representations through a new distillation framework to mitigate the domain discrepancy in the context of \textbf{V2X} 3D object detection. \textit{DI-V2X} comprises three essential components: a domain-mixing instance augmentation (DMA) module, a progressive domain-invariant distillation (PDD) module, and a domain-adaptive fusion (DAF) module. Specifically, DMA builds a domain-mixing 3D instance bank for the teacher and student models during training, resulting in aligned data representation. Next, PDD encourages the student models from different domains to gradually learn a domain-invariant feature representation towards the teacher, where the overlapping regions between agents are employed as guidance to facilitate the distillation process. Furthermore, DAF closes the domain gap between the students by incorporating calibration-aware domain-adaptive attention. 
Extensive experiments on the challenging DAIR-V2X and V2XSet benchmark datasets demonstrate \textit{DI-V2X} achieves remarkable performance, outperforming all the previous V2X models. Code is available at https://github.com/Serenos/DI-V2X.

\end{abstract}

\begin{figure}[t]
\centering
\includegraphics[width=0.45\textwidth]{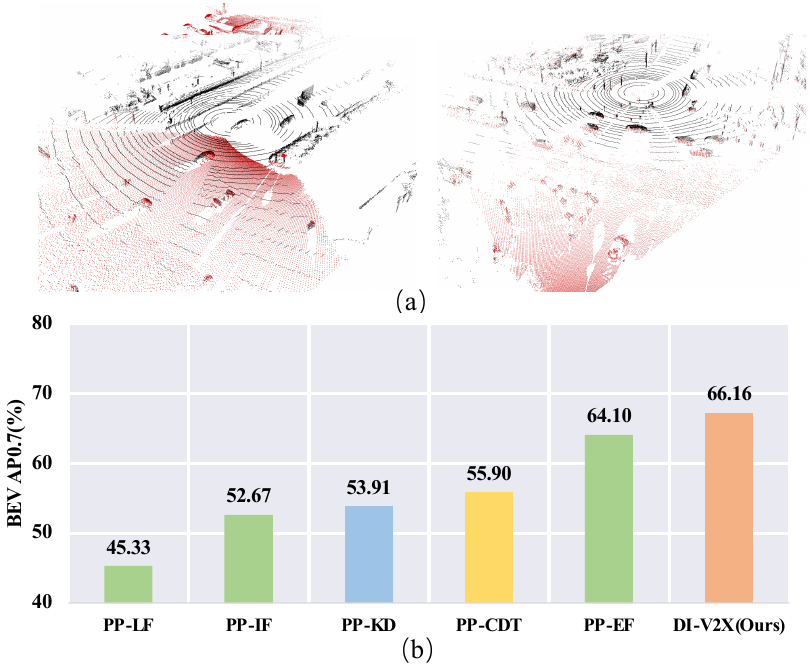}
\caption{(a) \textbf{Examples to show the data-level domain gap} caused by the different types of LiDAR sensors from the vehicle side (40-beam mechanical LiDAR colored in black) and the infrastructure side (300-beam solid-state LiDAR colored in red).
(b) \textbf{Performance comparison of different solutions} to mitigate the domain discrepancy, where our model attains superior performance.
}
\label{fig:top}
\end{figure}

\section{Introduction}
Vehicle-infrastructure collaborative perception~\cite{hu2022where2comm, xu2023bridging, lu2023robust}, also known as Vehicle-to-Everything (V2X) communication, has recently emerged as a promising approach to achieve full automation. It adequately leverages the sensor data collected from diverse agents, \ie, the vehicles and road-side infrastructures, to precisely perceive the complex driving scene. For instance, in situations where a vehicle's line of sight might be obstructed, information from infrastructures can serve as crucial redundancy due to their distinct viewpoints.  This collaboration essentially broadens the perception range and reduces occurrences of blind spots, enhancing the overall perception capability in contrast to previous single-vehicle autonomous driving systems.

To effectively fuse the information from various agents, leading V2X approaches tend to employ feature-based intermediate collaboration, which is referred to as intermediate fusion~\cite{wang2020v2vnet}. This approach retains essential information from each agent at the feature level, which can then be compressed for efficiency.
Therefore, intermediate fusion ensures a performance-bandwidth trade-off that is superior to the early fusion or late fusion approaches, where the former needs to transmit raw point cloud data between agents while the latter is vulnerable to incomplete results produced by each model. 

Nevertheless, current intermediate fusion models primarily concentrate on enhancing interaction between features from different agents. To date of this work, limited effort has been made to handle the data-level domain gap in the context of V2X collaborative perception. As depicted in Fig.~\ref{fig:top}(a), the vehicles and infrastructures might have different types of LiDAR sensors, thereby directly fusing the point cloud data or intermediate features from various sources could inevitably corrupt the final performance. 
Therefore, how to explicitly learn domain-invariant representation from multi-source data in this context remains to be explored. 

In this work, we propose a \textbf{\textit{D}}omain-\textbf{\textit{I}}nvariant \textbf{\textit{V2X}} model, \textbf{\ourmodel}, for collaborative 3D object detection in autonomous driving scenarios.  
To achieve this, a new teacher-student distillation model is introduced in \ourmodel. During training, we enforce the student models, \ie, the vehicles and infrastructures, to learn a domain-invariant representation aligned with that of the early-fused teacher model, \ie, the point clouds from multiple viewpoints are integrated into a holistic view to train the teacher. During inference, only the student models are retained. To be specific, \ourmodel~consists of three necessary components: a domain-mixing instance augmentation (DMA) module, a progressive domain-invariant distillation (PDD) module, and a domain-adaptive fusion (DAF) module. The purpose of DMA is to build a mixed ground-truth instance bank during training to align the point cloud inputs of the teacher and students, where instances are drawn from vehicle-side, road-side, or fused-side. 
After that, PDD aims to progressively transfer the information from the teacher to students across different stages, \ie, both before and after domain-adaptive fusion. 
For example, before fusion, students are guided to learn the domain-invariant representation separately in non-overlapping areas. While after fusion, we focus on the distillation within the overlapping area since the information has been well aggregated. In the DAF module, features from different domains are adaptively fused, based on their spatial significance. Additionally, DAF enhances the model's resilience to pose errors by integrating calibration offsets, ensuring robust V2X detection performance.

As illustrated in Fig.~\ref{fig:top}(b), we test different solutions that explicitly tackle the domain gap. `PP-LF', `PP-IF', and `PP-EF' indicate that we apply late fusion, intermediate and early fusion based on PointPillar~(PP)~\cite{Pointpillar}. Knowledge distillation (KD)~\cite{li2021learning} is an intuitive solution that adopts an early-fused teacher to guide the intermediate fusion process, while CDT refers to the cross-domain transformer in MPDA~\cite{xu2023bridging} to mitigate domain gap. All these solutions yield sub-optimal results compared to the proposed \ourmodel~model. 

 


To the best of our knowledge, \ourmodel~makes the early effort to bridge the data-level domain gap caused by sensor differences in the context of V2X collaborative perception, which is achieved by incorporating a domain-invariant distillation framework. The proposed modules PDD, DAF, and DMA prove highly effective in aligning, merging, and augmenting feature representations, ensuring the model acquires domain-invariant capabilities. Furthermore, we demonstrate the leading performance of \ourmodel~through extensive experiments on the challenging DAIR-V2X~\cite{yu2022dair} and V2XSet~\cite{xu2022v2x} benchmark datasets. It significantly outperforms previous state-of-the-art (SOTA) models like CoAlign~\cite{lu2023robust} and V2X-ViT~\cite{xu2022v2x} by 5.76\% AP@0.7 and 11.5\% AP@0.7, respectively. 


\section{Related Works}
\subsection{V2X Collaborative Perception}
V2X collaborative perception approaches can be roughly divided into three categories: early fusion, intermediate fusion, and late fusion. 
Early fusion models~\cite{chen2019cooper, yu2022dair} integrate the information from various agents at the raw data level, \ie, aggregating the 3D point cloud from multiple viewpoints to yield a holistic view of the scene.
However, transmitting the raw 3D data between agents inevitably consumes more communication bandwidth, which hinders its application in real-world scenarios.
late fusion methods~\cite{xu2023model, dair-seq}, on the other hand, performs fusion at the output level, \ie, the 3D bounding boxes. While these approaches can decrease the communication cost by only transmitting the perception results, the performance is susceptible due to the presence of noisy and incomplete outputs from each agent, as well as pose errors between agents.
To achieve a performance-bandwidth trade-off, intermediate-fusion methods~\cite{who2comm, when2comm, fcooper} are becoming popular as they conduct fusion at the feature level. They only transmit compressed feature maps that contain crucial information across agents, thereby significantly saving the communication cost while also remaining robust to pose errors in contrast to early fusion and late fusion models. 

In this paper, we mainly investigate the intermediate-fusion approaches. V2VNet~\cite{wang2020v2vnet} adopts a graph neural network to model the agent relation and aggregate messages from different vehicles. OPV2V~\cite{xu2022opv2v} utilizes self-attention to learn interactions between features. DiscoNet~\cite{li2021learning} leverages an early-fused teacher to guide the cooperation across different student agents. Where2comm~\cite{hu2022where2comm} focuses on the compression of feature maps, which only keep spatially sparse yet perceptually critical information. CoAlgin~\cite{lu2023robust} proposes an agent-object graph optimization method to handle the pose error. V2X-ViT~\cite{xu2022v2x} uses a transformer-based structure and a heterogeneous multi-agent self-attention mechanism to facilitate feature fusion.
 
All these approaches focus on enhancing interactions between the features from different agents or reducing communication costs while paying less attention to the inherent domain gap. 
To this end, we introduce \ourmodel~to explicitly learn domain-invariant representation to overcome the data-level domain disparities, which is achieved through a progressive distillation framework.

\begin{figure*}[t]
\centering
\includegraphics[width=0.8\textwidth]{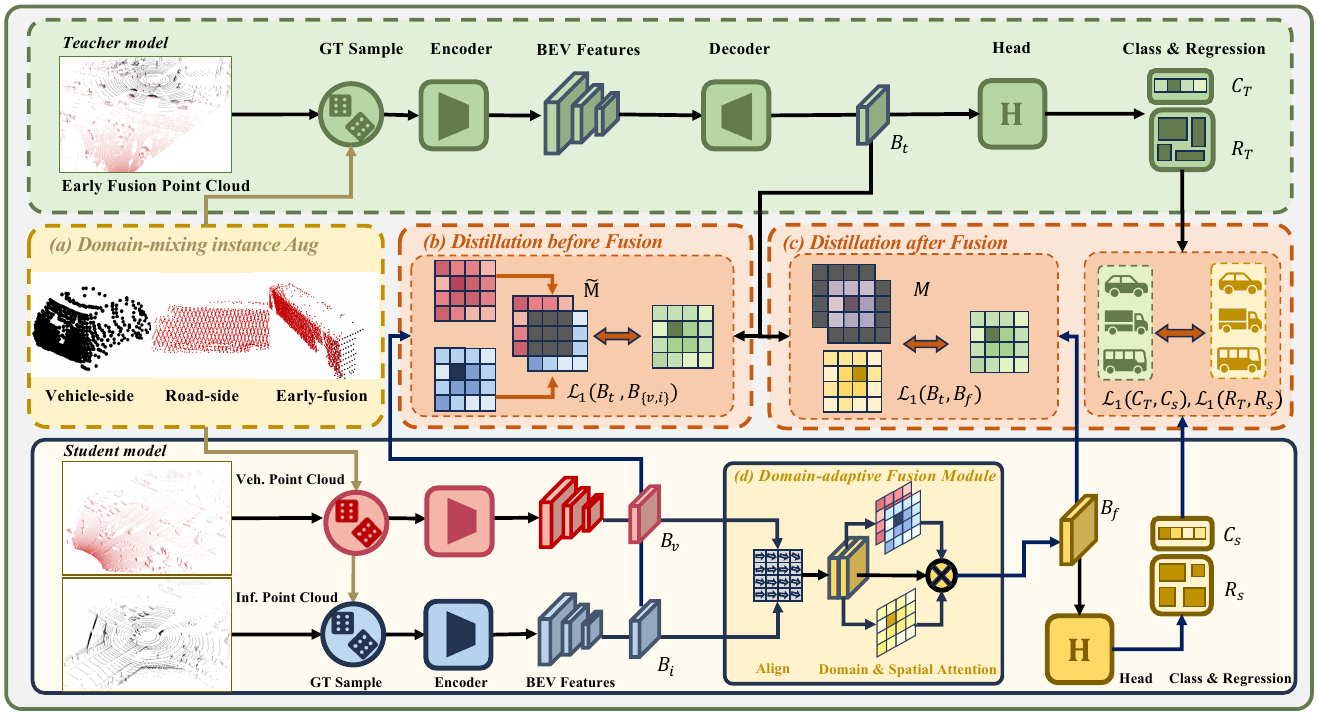}

\caption{\textbf{Overall architecture of \ourmodel.} It consists of a teacher model that takes the holistic-view point cloud as input, along with two student models utilizing point clouds from different domains, \eg, the vehicles and infrastructures. \textbf{(a).} A Domain-Mixing instance Augmentation (DMA) module is used to generate aligned data representation. \textbf{(b-c).} A Progressive Domain-invariant Distillation (PDD) module then aligns the students with the teacher to learn domain-invariant features, both before and after fusion with the guidance of overlap information. \textbf{(d).} A Domain-Adaptive Fusion (DAF) module further enhances the feature with domain- and spatially-adaptive attention. The components within the dashed boxes are only used during training.
}
\label{fig:framework}
\end{figure*}

\subsection{Domain-invariant Representation in Point Cloud}
In the context of point cloud-based perception, domain gaps may arise from variations in sensor types, weather conditions, and geographical factors. Domain-invariant learning~\cite{survey_da_lidar} is a promising approach to ensure feature generalization in V2X perception tasks. There are typically two approaches to achieve this: constructing domain-invariant data representation and conducting domain-invariant feature learning. 
To attain domain-invariant data, 
some works tend to apply data augmentation strategies~\cite{PointMixup, PointCutMix, Mix3D} to transfer the multi-source data into a unified domain. LiDomAug~\cite{InstanceAug_SemanticSeg} proposes a sophisticated framework to generate a dense point cloud scene with sequential frames to adapt to random configurations and reduce sensor bias.
Other works apply scene-level augmentation~\cite{alonso2021domain} or re-sample the LiDAR scans to match the data distribution of target domain~\cite{DA_SS}.
On the other hand, some works focus on domain-invariant representation learning to create representations that capture underlying patterns and characteristics common across different domains. 

Most of the aforementioned approaches target a domain adaptation task that seeks to transfer the knowledge from a source domain to a target domain. In contrast, we focus on acquiring a generalized representation capable of unifying point clouds from diverse domains (\textit{e.g.,} vehicle-side and road-side) for improved performance in V2X perception.

\section{The Proposed \ourmodel~Framework}

In typical driving scenarios with V2X collaborative perception, vehicles and infrastructures are commonly equipped with different types of LiDAR sensors,
which will result in data-level domain gap. 
In this work, we aim to tackle the V2X collaborative 3D object detection task from the perspective of domain-invariant feature learning. In this section, we will introduce the overall framework. Next, we will delve into the details of each module. 

\subsection{Overall Architecture}

As illustrated in Fig.~\ref{fig:framework}, \ourmodel~adopts a teacher-student distillation framework that aims to learn domain-invariant representations for the student models. Formally, let $\mathbf{P}_{v} \in \mathbb{R}^{N_{v}\times 4}$ and $\mathbf{P}_{i} \in \mathbb{R}^{N_{i}\times 4}$ denote the vehicle-side and road-side point clouds which have been temporally synchronized. We first fuse the two point clouds from distinct viewpoints to attain a holistic viewpoint cloud $\mathbf{P}_{e}$ by leveraging their pose information. Then, $\mathbf{P}_{e}$ is utilized to train the teacher.

To be specific, we employ the Domain-Mixing instance Augmentation (DMA) module to get augmented inputs $\{\mathbf{P}_{v}^{'}, \mathbf{P}_{i}^{'}, \mathbf{P}_{e}^{'}\} = Aug(\{\mathbf{P}_{v}, \mathbf{P}_{i}, \mathbf{P}_{e}\})$. The perception model is instantiated by a PointPillars~\cite{Pointpillar} 3D detector here.
We discretize the $\mathbf{P}_{e}^{'}$ into voxels and employ a 3D encoder (\eg, VoxelNet~\cite{Second}) to extract bird-eye-view (BEV) feature maps denoted as $\mathbf{B}_{t} \in \mathbb{R}^{H\times W \times C}$. 
Finally, a detection head predicts the class and regression results $\mathbf{C}_{T} = \{\mathbf{c}_{k}\}_{k=1}^{K}, \mathbf{R}_{T} = \{\mathbf{r}_{k}\}_{k=1}^{K}$. $\mathbf{r}_{k} = (x,y,z,h,w,l,\theta)$, where $(x,y,z)$ and $(h,w,k)$ represent the center coordinate and size of an object and $\theta$ indicates the orientation along the $X$-$Y$ plane. 

The student models share the most designs with the teacher model, \eg, voxel configuration, 3D encoder, and detection head. The augmented point clouds $\{\mathbf{P}_{v}^{'}, \mathbf{P}_{i}^{'}\}$ are sent to a share-weight 3D encoder to extract BEV features $\mathbf{B}_{v}, \mathbf{B}_{i} \in \mathbb{R}^{H\times W \times C}$, respectively. Then, a Domain-adaptive Fusion (DAF) module is presented to fuse the feature maps from different domains to get $\mathbf{B}_{f} \in \mathbb{R}^{H\times W \times C}$. 
During training, the proposed Progressive Domain-invariant Distillation (PDD) module aligns various student features ($\mathbf{B}_{v}$, $\mathbf{B}_{i}$ and $\mathbf{B}_{f}$) with the teacher BEV feature ($\mathbf{B}_{t}$) both before and after fusion, which is guided by the overlapping and non-overlapping mask $\mathbf{M}, \mathbf{\tilde{M}} \in \{0,1\}^{H\times W}$. In this way, the students can learn a domain-invariant representation that is adaptable for multi-source data in the V2X context.
After training, only the student is employed for inference 

\subsection{Domain-mixing Instance Augmentation}
For 3D perception in autonomous driving scenarios, instances are more informative than the noisy background. Therefore, we propose to augment the input point cloud with domain-mixing instances, aligning the data distribution between the teacher and students.
This also facilitates the subsequent distillation process. Specifically, let $\mathbf{P}_{v} \in \mathbb{R}^{N_{v}\times 4}, \mathbf{P}_{i} \in \mathbb{R}^{N_{i}\times 4}$ be the point clouds from vehicle and infrastructure. We first project $\mathbf{P}_{i}$ to the coordinate system of ego-vehicle, such that $\mathbf{P}_{i}^{T} = \mathbf{T}_{(i \rightarrow v)}\mathbf{P}_{i}^{T}$, where $\mathbf{T}_{(i\rightarrow v)} \in \mathbb{R}^{4\times 4}$ is the transformation matrix from infrastructure to vehicle system. Then we leverage the ground-truth bounding box $\mathbf{B}_{gt} = \{\mathbf{b}_{k}\}$ to get instances from $\mathbf{P}_{v}$ and $\mathbf{P}_{i}$.
Instance from different domains corresponding to the same ground-truth object will be merged to get an early fusion instance $\mathbf{p}_{k} = \text{Concat}(\mathbf{p}_{k}^{v}, \mathbf{p}_{k}^{i}) \in \mathbb{R}^{N_{k}\times 4}$, where $\mathbf{p}_{k}^{v}\in\mathbf{P}_{v}$ and $\mathbf{p}_{k}^{i}\in\mathbf{P}_{i}$ are instance points indexed by $\mathbf{b}_{k}$ from both domains. As relative positions between agents change dynamically with the motion of the ego-vehicle, some instances may only come from individual domains, while some others may come right from the overlapping area that has been early fused. To determine the domain source of each instance, we divide these instances into three categories by computing the point proportion from each domain:
\begin{equation}
    \begin{aligned}
    D_{i} &= \{\mathbf{p}_{k} | N_{k}^{v}/(N_{k}^{v} + N_{k}^{i}) < \tau_{l}\} \\
    D_{f} &= \{\mathbf{p}_{k} | \tau_{l} < N_{k}^{v}/(N_{k}^{v} + N_{k}^{i}) < \tau_{h}\} \\
    D_{v} &= \{\mathbf{p}_{k} | N_{k}^{v}/(N_{k}^{v} + N_{k}^{i}) > \tau_{h} \}
\end{aligned}
\end{equation}
where $N_{k}^{v}$ and $N_{k}^{i}$ represent the point number from the vehicle side and infrastructure side, respectively, and $\tau_{l}, \tau_{h}$ indicate the thresholds.  
Then, we can get an instance bank $D_{mixed} = D_{i} \cup D_{f} \cup D_{v}$, which contains mixed instances from all the domains, \ie, including the fused domain.
During training, we randomly sample instances from $D_{mixed}$ according to certain probability and add these instances to both the teacher and students.

By involving instances from different domains, we enhance the diversity of the training data. Furthermore, from each student's perspective, information from other domains is incorporated through instance-level mixup~\cite{Mixup}. This approach essentially aligns the data distributions between the teacher and student models, thereby resulting in more generalized features after the subsequent knowledge distillation process.

\subsection{Progressive Domain-invariant Distillation}
In this section, we focus on acquiring domain-invariant features through progressive domain-invariant distillation. To achieve this, 
We adopt a two-stage distillation strategy, \ie, distillation before and after the domain-adaptive fusion (DAF) module. The first distillation stage is to align the students' distributions with the teacher model as inputs for DAF, which is essential for accurate information fusion. However, we empirically found that directly conducting distillation over the whole feature map between students and the teacher yields suboptimal performance. 
To address this, we opt for distillation in the non-overlapping area during the first stage. In the second stage, since the student features have been well fused by DAF, we can concentrate on distillation over the overlapping area. This two-stage distillation process enforces the student models to match the feature representations of the teacher model from different regions, reducing the gap between students.
\\
\noindent\textbf{Distillation Before Fusion}. 
To perform the first-stage distillation, we need to first compute the overlapping mask to determine the overlapping area. 
We present the vehicle perception region as a rectangle $\mathbf{A}_{v} = (x_{v}, y_{v}, 2R_{x}, 2R_{y}, \theta_{v})$, where $R_{x}, R_{y}$ are the max range along $X$-$Y$-plane and $\theta$ denotes the rotation. All the computations are unified under the vehicle coordinate system, thus $(x_{v}, y_{v})=(0, 0)$ and $\theta_{v}=0$. The perception region of the infrastructure side is then transformed to the vehicle side, and get a new rectangle $\mathbf{A}_{i} = (x_i, y_i, 2R_{x}, 2R_{y}, \theta_{i})$. Then we can compute the overlapping area between $A_{v}$ and $ A_{i}$, which is $\mathbf{P}_{overlap}=Intersection(\mathbf{A}_{v}, \mathbf{A}_{i})$. The obtained $\mathbf{P}_{overlap}$ (\ie, a polygon) is then downsampled to match the size of feature map $\mathbf{B}_v \in \mathbb{R}^{H \times W \times C}$. The overlapping mask $\mathbf{M}\in \{0,1\}^{H\times W}$ is then defined by:
\begin{align}
    \mathbf{M}(i,j) = \left\{
    \begin{array}{ccc}
    1&,   & \text{if} \ (i,j)\in \mathbf{P}_{overlap}     \\
    0&,   & \text{otherwise}    \\
    \end{array}
    \right.
\end{align}
 where $\mathbf{M}(i,j) \in {0,1}$ denotes the binary value at the coordinate of $(i,j)$.
By focusing distillation only on the non-overlapping area $\mathbf{\tilde{M}}$, we allow each student to concentrate on learning representations that align their respective domains. This avoids strict constraint that enforces incomplete student features to learn towards complete features from the teacher. The distillation loss is then expressed by:
\begin{equation}
\begin{aligned}
  \mathcal{L}_{da} & = \mathcal{L}_{1}(\mathbf{B}_{t}, \mathbf{B}_{v}\odot \mathbf{\tilde{M}}_{v}) + \mathcal{L}_{1}(\mathbf{B}_{t}, \mathbf{B}_{i}\odot \mathbf{\tilde{M}}_{i}) \\
  & = \frac{1}{HW}\sum_{m}^{H} \sum_{m}^{W} |\mathbf{B}_{t}(m,n)-\mathbf{B}_{v}(m,n)| \times \mathbf{\tilde{M}}_{v}(m,n) \\
  & +\frac{1}{HW}\sum_{m}^{H} \sum_{n}^{W} |\mathbf{B}_{t}(m,n)-\mathbf{B}_{i}(m,n)| \times \mathbf{\tilde{M}}_{i}(m,n)
\end{aligned}
\end{equation}
where $\mathcal{L}_{1}$ is the L1 loss and $\odot$ denotes the element-wise multiplication with channel-wise broadcasting.$ \tilde{M}_{v},\tilde{M}_{i}$ are the non-overlapping mask for vehicle and infrastructure.\\
\noindent\textbf{Distillation After Fusion}. In this phase, the student features from various domains have been effectively merged using the DAF module. As a result, we obtain a highly capable fused representation denoted as $\mathbf{B}_{f}$, which can be aligned with the teacher's feature representation $\mathbf{B}_{t}$. Intuitively, $\mathbf{B}_{t}$ is obtained through the early collaboration of mixed point cloud data, which inherently involves minimal information loss. By enforcing the intermediate fused features $\mathbf{B}_{f}$ to progressively align with $\mathbf{B}_{t}$, we effectively ensure that the essential knowledge gained through the early fusion stage is consistently integrated throughout the learning process, resulting in the domain-invariant feature representations. Formally, the second-stage distillation loss can be denoted as:
\begin{equation}
\begin{aligned}
  \mathcal{L}_{f} & = \mathcal{L}_{1}(\mathbf{B}_{t}, \mathbf{B}_{f}\odot \mathbf{M}_{v}) \\
  & = \frac{1}{HW}\sum_{m}^{H} \sum_{m}^{W} |\mathbf{B}_{t}(m,n)-\mathbf{B}_{f}(m,n)| \times \mathbf{\tilde{M}}_{v}(m,n) \\
\end{aligned}
\end{equation}
Moreover, we could go beyond the feature level alignment and extend to the prediction level. Since our final target is to decode the final 3D bounding boxes from both $\mathbf{B}_{f}$, ensuring alignment at the prediction level further contributes to the consistency and accuracy of the results. The prediction-level distillation can be formulated as:
\begin{equation}
\begin{aligned}
  \mathcal{L}_{p} & = \mathcal{L}_{class} + \mathcal{L}_{regression} \\
  & = \frac{1}{K}\sum_{k}^{K}(|\mathbf{c}_{k}-\mathbf{c}^{s}_{k}|+|\mathbf{r}_{k}-\mathbf{r}^{s}_{k}|)\\
\end{aligned}
\end{equation}
where $K$ is the number of predictions, $\mathbf{C}_T = \{\mathbf{c}_{k}\}_{k=1}^{K}, \mathbf{R}_T = \{\mathbf{r}_{k}\}_{k=1}^{K}$ indicate the class and regression prediction of teacher model and $\mathbf{C}_{S} = \{\mathbf{c}_{k}^{s}\}_{k=1}^{K}, \mathbf{R}_{S} = \{\mathbf{r}_{k}^{s}\}_{k=1}^{K}$ refer to the class and regression prediction of student model after feature fusion.
During training, the overall loss function is
\begin{align}
    \mathcal{L} = \mathcal{L}_{detect} + \lambda_{kd}(\mathcal{L}_{da} + \mathcal{L}_{f} + \mathcal{L}_{p})
\end{align}
where $\mathcal{L}_{detect}$ is the detection loss. 
%
%
%
\begin{figure}[t]
\centering
\includegraphics[width=0.45\textwidth]{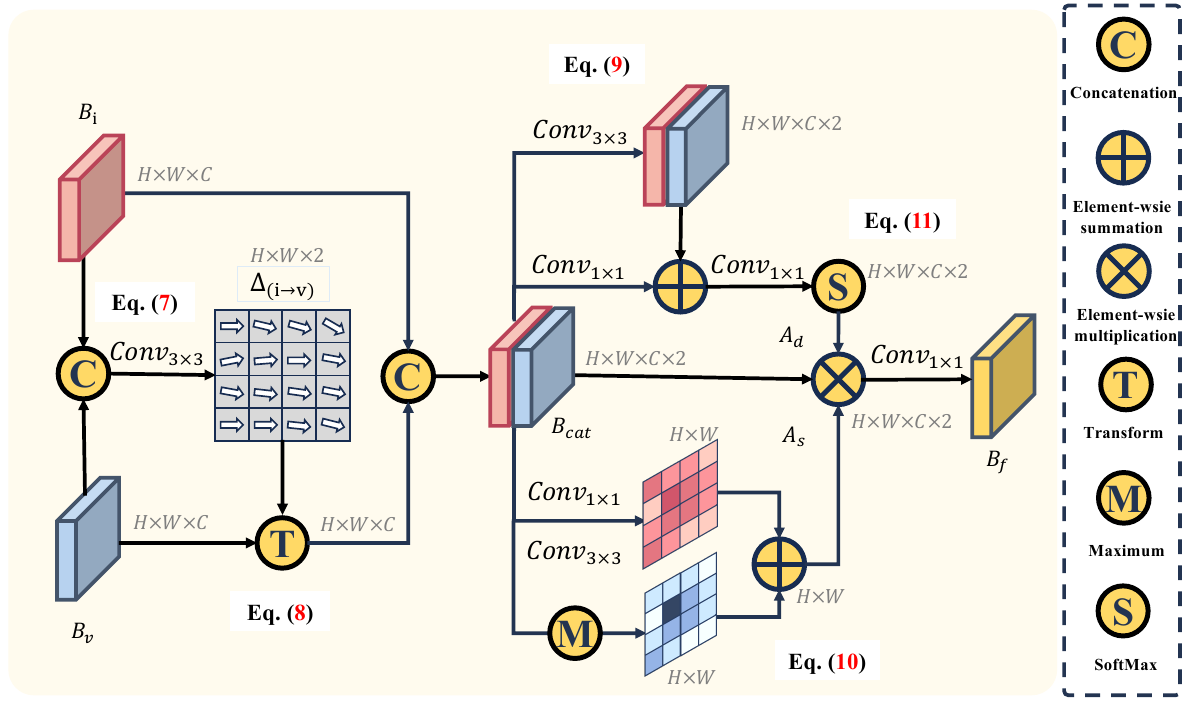}
\caption{\textbf{Illustration of the Domain-Adaptive Fusion module.} Features from different domains are first aligned by a collaboration offset to reduce pose errors. Then, the aligned features are further refined by domain-adaptive and spatially-adaptive attention mechanisms.
}
\label{fig:dgfm}
\end{figure}

\begin{table*}[t]
    \centering\small
        \scalebox{1.0}{
        \begin{tabular}{c|c|c|c|c}
            \hline
            {Fusion} & {Model} & Publication & BEV AP@0.5 & BEV AP@0.7\\ 
            \hline	 \hline
            \textit{No Fusion} & PP~\cite{Pointpillar}
            & CVPR 2019 & 65.21 & 54.26\\
            \hline
            \textit{Late Fusion} & PP-LF & CVPR2019 & 63.68 & 45.33\\
            \hline
            \multirow{2}{*}{\textit{Early Fusion}}
                & PP-EF & CVPR2019 & 76.55& 64.10\\
                & \ourmodel~(teacher) & - & \textbf{79.81}& \textbf{67.08}\\
            \hline \hline

            \multirow{8}{*}{\textit{Intermediate Fusion}}
                & PP-IF & CVPR 2019 &72.85 & 52.67 \\
                & V2VNet~\cite{wang2020v2vnet}& ECCV 2020 & 70.97 & 47.63\\
                & DiscoNet~\cite{li2021learning}& NeurIPS 2021 & 73.87& 59.61\\
                & OPV2V~\cite{xu2022opv2v}& ICRA 2022 & 73.30&55.33\\
                & V2X-ViT~\cite{xu2022v2x}& ECCV 2022 & 71.81 & 54.94\\
                & Where2comm~\cite{hu2022where2comm} & NeurIPS 2022 & 74.09& 59.66\\
                & CoAlign~\cite{lu2023robust}& ICRA 2023 & \underline{74.60}& \underline{60.40}\\
                \cline{2-5}
                & \ourmodel~(student) & - & \textbf{78.82}(\color[rgb]{0,0.4,0}$+4.22$)& \textbf{66.16}(\color[rgb]{0,0.4,0}$+5.76$)\\
            \hline	
        \end{tabular}
        }
    \caption{\textbf{Comparison with state-of-the-art methods} on DAIR-V2X \texttt{val} dataset. 
    We apply PointPillars (denoted as PP) as the detection model to benchmark different fusion approaches (\ie, no/early/intermediate/late fusion).
    \ourmodel~(teacher) refers to the early fused teacher trained with DMA strategy, while \ourmodel~(student) is the intermediate fusion model. The best and the second-best approaches in each column are indicated by bold and underlined formatting.
    }
    \label{table:result-main}
\end{table*}

\subsection{Domain-Adaptive Fusion} 
The target of the DAF module is to aggregate the features from both vehicles and infrastructure, creating an enhanced representation that contains valuable information from each domain. However, this fusion process comes with two main challenges: potential misalignment of the pose between two sides and the design of a suitable feature interaction strategy. {The relative pose of vehicles and infrastructure in the real world is vulnerable due to facts such as sensor noise, dynamic motion, and inconsistency in varying timestamps, which will impact the accuracy of V2X perception. To address this, we proposed to leverage the calibration offset to dynamically correct the potential pose errors. Regarding feature collaboration, our approach emphasizes essential domain-specific information across  spatial locations.}

The overview of DAF is illustrated in Fig~\ref{fig:dgfm}. We first predict the calibration offset with a convolution layer to better align $\mathbf{B}_{i}$ with $\mathbf{B}_{v}$, which is denoted as:
\begin{align}
    \mathbf{\Delta}_{(i\rightarrow v)} = \text{Conv}(\text{Concat}(\mathbf{B}_{v}, \mathbf{B}_{i})) \in \mathbb{R}^{H \times W \times 2}.   
\end{align}
Then we can use this offset to transform the infrastructure feature $\mathbf{B}_{i}$ to align with the vehicle feature $\mathbf{B}_{v}$.
\begin{align}
   \mathbf{B}_{i}^{'}(p_{k}) = \mathbf{B}_{i}(p_{k} + \mathbf{\Delta}_{(i\rightarrow v)}(p_{k})), 0\le k < HW 
\end{align}
where $p_{k} \in \mathbb{R}^{2}$ indicates the position of $k$-th grid and $\mathbf{B}(p)$ means the feature value at position $p$. This operation enables the dynamic integration of information from neighboring points, thereby mitigating potential pose errors.

Furthermore, we aim to extract both domain-adaptive and spatially-adaptive features from $\mathbf{B}_{i}$ and $\mathbf{B}_{v}$. This allows the model to leverage both the unique information present in each domain and the shared spatial context, enhancing the overall representation. Mathematically, we first aggregate feature maps by concatenation operation to get $\mathbf{B}_{cat} = \text{Concat}(\mathbf{B}_{v}, \mathbf{B}{i}^{'}) \in \mathbb{R}^{H \times W \times C \times 2}$. Domain-adaptive attention can be formulated as:
\begin{align}
    \mathbf{A}_{d} = \text{Softmax}(\text{Conv}(\mathbf{B}_{cat})) \in \mathbb{R}^{H \times W \times C \times 2}
\end{align}
where $\text{Conv}(\cdot)$ is implemented by a $3\times 3$ convolution layer and two $1\times 1$ convolution layers with residual structure. $\text{Softmax}(\cdot)$ is used to generate the weight for each domain. 
The domain-adaptive attention can dynamically adjust the weight of each domain in each region, resulting in a more flexible collaboration.
Spatial attention is also crucial because it forces the network to focus on the areas with high objectness scores or areas that can provide essential context information. Here, we use convolution of various kernel sizes to aggregate features with different receptive fields and a $Maximum$ function to select important areas. The spatially-adaptive attention can be formulated as:
\begin{align} 
    \mathbf{A}_{s} = \text{Conv}(\mathbf{B}_{cat}) + \text{max}(\mathbf{B}_{cat})
\end{align}
where $\mathbf{A}_{s} \in \mathbb{R}^{H \times W} $, $\text{max}(\cdot)$ is the maximum value along the channel and domain dimension and $\text{Conv}(\cdot)$ is two convolution layers with different kernel size. Finally, we get the fused feature considering both domain and spatial attention:
\begin{align}
    \mathbf{B}_{f} = \text{Conv}(\mathbf{A}_{d}\odot \mathbf{A}_{s} \cdot \mathbf{B}_{cat})  \in \mathbb{R}^{H \times W \times C}
\end{align}
where the $\text{Conv}$ is the convolution layer to reduce channels, $\odot$ means element-wise multiplication with channel wise broadcast and $\cdot$ is the element-wise multiplication.  
Thus, spatially-adaptive attention can provide a robust and flexible attention map by aggregating multi-granularity features.
\begin{table}[t]
		\centering\small
		\scalebox{0.9}{
    		\begin{tabular}{c|c|c|c}
    			\hline
                    Fusion Paradigm& Model& AP@0.5& AP@0.7\\
                    \hline \hline
                    \textit{No Fusion}    & \multirow{3}{*}{PointPillar(PP)} & 60.6 & 40.2\\
                    \textit{Late Fusion}  & & 72.7& 62.0\\
                    \textit{Early Fusion} & & 81.9& 71.9\\
                    \hline
                    \multirow{5}{*}{\textit{Intermediate} Fusion} 
                    & PP-IF    & 76.9 & 49.2\\
                    & DiscoNet    & 84.4 & 69.5\\
                    & Where2comm  & 85.5 & 65.4\\
                    & V2X-ViT     & \underline{88.2} & \underline{71.2}\\
                    & \textit{DI-V2X} (student)      & \textbf{92.7}(\color[rgb]{0,0.4,0}$+4.5$) & \textbf{82.7}(\color[rgb]{0,0.4,0}$+11.5$)\\
    			\hline	
    		\end{tabular}
		}
		\caption{\textbf{Performance comparasion} on V2XSet dataset.
        }
		\label{table:result-v2xset}
	\end{table}

\section{Experiments}
\subsection{Dataset}
\textbf{DAIR-V2X}. We employ the challenging DAIR-V2X~\cite{yu2022dair} for evaluating our model and other SOTA approaches.
DAIR-V2X is a real-world V2X collaborative perception dataset, which is collected by well-equipped vehicles and infrastructures. The road-side infrastructures are equipped with 300-beam solid-state LiDAR with $100^{\circ}$ horizontal FOV, while the vehicle uses 40-beam mechanical LiDAR with $360^{\circ}$ horizontal FOV.  
Totally 9K synchronized vehicle and infrastructure LiDAR frame pairs are sampled from 100 representative scenes at 10Hz. DAIR-V2X provides 3D object detection labels within the camera's field of view. We utilize these comprehensive 3D labels that encompass a 360-degree detection range, as mentioned by prior works~\cite{lu2023robust}.
\\
\textbf{V2XSet}. We also evaluate our method on another V2X calibration dataset V2XSet~\cite{xu2022v2x}, which contains 6694 training data and 1920 validation data generated by the simulator. Each scene contains point clouds collected from at least 2 and at most 7 agents. These agents are equipped with 36-beam LIDAR with $360^{\circ}$ horizontal FOV.

\subsection{Experimental setup}

\textbf{Evaluation metrics}. V2X collaborative 3D object detection aims to achieve both high detection performance and minimal communication costs. Our focus is on enhancing the detection performance without increasing the communication bandwidth compared with other intermediate fusion approaches. AP~(average precision) is employed to evaluate the detection performance in BEV at Intersection-over-Union~(IoU) thresholds of 0.5 and 0.7.
\\
\textbf{Implementation Details}. We set the point cloud range to $[-100, 100]\times [-40, 40]\times [-3.5, 1.5]$ meters defined in the vehicle coordinate system with the voxel size as $[0.4, 0.4, 5]$ meters along $XYZ$ axes. For scene-level data augmentation, we adopt random flip, rotation, and scaling. 
The proposed DMA randomly samples the instances from the database with the probability of 0.8, 0.1, and 0.1 for $D_f, D_v, D_i$. We adopt the PointPillars~\cite{Pointpillar} as the detector to align with other V2X approaches~\cite{lu2023robust}, while our approach is versatile to any BEV-based 3D object detectors. The feature map after fusion (\ie,$B_{i}, B_{v}$) is of size $252\times 100 \times 256$. We set $\lambda_{kd}$ as $1$ and the thresholds $\tau_{l},\tau_{h}$ as 0.2 and 0.8. All the models are trained on 4 NVIDIA Tesla V100 GPUs with a batch size of $4$ for $40$ epochs. 

\subsection{Main Results}
\label{sec:exp}
\textbf{Performance Benchmarking.}
Table~\ref{table:result-main} presents a comparison of various V2X approaches, including early, intermediate, and late fusion methods. {PP} refers to PointPillars co-trained with mixed data of vehicle and infrastructure.  
As presented in Table~\ref{table:result-main}, we can observe that the late fusion method {PP-LF}, which directly combines the detection results from two sides using non-maximum suppression (NMS) even results in a performance decrease of 1.53\% in terms of AP@0.5, probably due to the pose error. The naive intermediate fusion method {PP-IF}, which is obtained by directly summing the features from both domains, improves the performance by 7.64$\%$ in terms of AP@0.5. Furthermore, the early fusion method {PP-EF} surpasses other fusion types by a large margin. By utilizing domain-mixing augmentation, our \ourmodel~(teacher) achieves the highest performance in terms of both BEV AP@0.5 and AP@0.7. 

Additionally, we conduct a comparison between our \ourmodel~and some competitive intermediate fusion methods.
The results of OPV2V and CoAlign are borrowed from~\cite{lu2023robust}. Besides, we reimplement V2VNet, V2X-ViT and DiscoNet and achieve slightly better performance than the results reported in CoAlign.
We can see that \ourmodel~(student) surpasses all these intermediate fusion methods. For example, it outperforms the previously best-performing model CoAlign by 4.22\% AP@0.5. This demonstrates the importance of domain-invariant learning.

In Table~\ref{table:result-v2xset}, we present performance comparisons on another challenging dataset, V2XSet~\cite{xu2022v2x}. We access different fusion strategies and compare our model with SOTA models like DiscoNet, Where2comm, and V2X-ViT. The results indicate that \ourmodel~outperforms these approaches significantly on the V2XSet validation set, demonstrating the generalizability of our model.
\\
%
%
\begin{table}[t]
\centering\small
    \scalebox{0.9}{
    \begin{tabular}{c|c|c|c|c|c}
        \hline
             \multirow{2}{*}{Model} & \multirow{2}{*}{Training} & \multicolumn{3}{c|}{Testing} & {Avg.}\\
            \cline{3-5}
            & & Veh. & Inf. &  Veh. $\&$ Inf.& AP@0.7\\
            \hline \hline
            \multirow{2}{*}{PP}& Veh.& {73.08}&31.53& 52.62& 52.41\\
            & Inf.& 32.21& \textbf{65.80}& 26.39 & 41.46\\
            \hline
            PP-LF & \multirow{5}{*}{Veh. $\&$ Inf.}& 63.62 & 65.66 & 45.33 &  58.20\\
            PP-EF & & 73.19 & 47.44 & 64.10 & 61.57\\
            PP-IF & & 68.55 & 7.45 & 52.67 & 26.22\\
            PP-KD & & 69.91 & 17.82 & 53.91 & 47.11\\
            PP-CDT & & 66.22 & 41.73 & 55.90 & 54.61\\
            \hline 
            \hline
            Ours-T & \multirow{2}{*}{Veh. $\&$ Inf.} & \textbf{76.74}& 53.34& \textbf{67.08} & \textbf{65.72}\\
            Ours-S & & 73.04 & 45.91 & 66.16 &  61.70 \\
        \hline	
    \end{tabular}
    }
    \caption{\textbf{Domain generalization ability of different solutions} on DAIR-V2X \texttt{val} set. `Veh.' and `Inf.' indicate the vehicle and infrastructure domain, and `Veh.$\&$ Inf.' means using data from both sides. `Avg.' is the average AP.} 
    \label{table:domain-generation}
\end{table}
\\
\textbf{Domain generalization ability of \ourmodel.}
As our method aims to learn domain-invariant representation, we analyze the effectiveness of \ourmodel~in improving the model generalization ability in this section. 
To this end, we evaluate the transfer learning results (\ie, training on Veh. $\&$ Inf. and testing on Veh., Inf, or Veh. $\&$ Inf.). 
We compare with some common solutions that mitigate domain gap, \eg, PP-LF, PP-EF, PP-IF, PP-KD, and PP-CDT. PP-KD adopts an early-fusion teacher to guide the collaboration through knowledge distillation while PP-CDT leverages a cross-domain transformer to align the domain distribution~\cite{xu2023bridging}. In cases of communication failure, vehicles may not receive signals from roadside infrastructure, \ie, testing only with vehicle data. Here, our model achieves competitive results with a model trained with only the vehicle data. Among the intermediate fusion approaches, our model also gives the best performance when testing only with the infrastructure data. When data from both sides are available, our model obtains the best performance. In essence, our model attains the best average accuracy across various cases, showing its superior generalization capability. \\

%
\begin{table}[t]
		\centering\small
		\scalebox{1.0}{
    		\begin{tabular}{c|c|c|c|c|c}
    			\hline
                    Model& DMA& PDD & DAF& AP@0.5&  AP@0.7\\
                    \hline \hline
                    PP-IF & -& -& -& 72.85& 52.67\\
                    \cline{1-6}
                    \multirow{5}{*}{DG-V2X}& \checkmark& -& -& 73.90&  54.02\\
                    & -& \checkmark& & 74.17& 55.89\\
                    & -& -& \checkmark& 75.76 & 60.89 \\
                    & -& \checkmark & \checkmark& 77.37 & 61.66  \\
                    & \checkmark& \checkmark& \checkmark& \textbf{78.82} & \textbf{66.16} \\
    			\hline	
    		\end{tabular}
		}
		\caption{\textbf{Ablation studies of \ourmodel} to verify each proposed module on the DAIR-V2X \texttt{val} set. Our final model surpasses the baseline fusion model, \ie, PP-IF, by 5.97$\%$ and 13.49$\%$ in terms of AP@0.5 and AP@0.7, respectively.
        }
		\label{table:result-ablation}
	\end{table}
 
\noindent\textbf{Ablation Study}. We investigate the contribution of each proposed module in this section, where an intermediate fusion model PP-IF is used as the baseline. PP-IF refers to PointPillars trained by summing features from both domains. As illustrated in Table~\ref{table:result-ablation}, the three proposed modules are all beneficial to the final performance. To be specific, 
%
%
DMA improves the AP@0.7 by $1.35$\% and AP@0.5 by $1.05$\%, which indicates that domain-mixing instances sampling for both agents enriches the data diversity and aligns the distribution.
DAF improves the AP@0.7 by $8.22$\% and AP@0.5 by $2.91$\% by implicitly reducing the sensor noise of the two agents and enriching their feature interaction. 
To evaluate PDD, we equip PP-IF with an early fused teacher model and distill features both before and after fusion guided by an overlapping mask and improve the AP@0.7 by $3.22\%$ and AP@0.5 $1.32\%$ 
The performance can be further boosted when applying three modules, which proves the effectiveness of ~\ourmodel.

\section{Conclusion}
In this work, we presented a novel approach, \textit{DI-V2X}, to address a critical challenge in V2X collaborative perception, \ie, data-level domain gap due to sensor differences across agents. The core is to leverage a domain-invariant distillation framework to enhance the generalization of the feature representation. To achieve that, a Domain-Mixing instance Augmentation (DMA) strategy is proposed to ensure the alignment of data distributions between teacher and student models. Then, a Progressive Domain-invariant Distillation (PDD) module leverages overlap information shared between agents to facilitate the distillation both before and after domain fusion. The Domain-Adaptive Fusion (DAF) module further aggregates features from each domain by considering both domain-adaptive and spatially relevant information. It also accounts for potential pose errors by incorporating calibration offsets. Experimental results conducted on the real-world datasets DAIR-V2X and V2XSet demonstrate the remarkable performance of our approach.

\newpage


\bibliography{aaai24}

\end{document}